\documentclass[10pt,twocolumn,letterpaper]{article}

\usepackage{cvpr}              

\definecolor{verylightgray}{rgb}{0.88, 0.88, 0.88} 
\definecolor{verylightblue}{rgb}{0.8, 0.9, 1.0} 
\definecolor{lightpurple}{rgb}{0.9, 0.9, 1.0}

\usepackage{multirow}
\usepackage{algorithm}
\usepackage{algorithmic}
\usepackage{pifont}  

\definecolor{cvprblue}{rgb}{0.21,0.49,0.74}
\usepackage[pagebackref,breaklinks,colorlinks,allcolors=cvprblue]{hyperref}

\def\paperID{1961} 
\def\confName{CVPR}
\def\confYear{2026}

\def\tracker{PNTrack}

\title{Boosting Self-Supervised Tracking with Contextual Prompts and Noise Learning}

\author{Yaozong Zheng\textsuperscript{\rm 1,2}, Qihua Liang\textsuperscript{\rm 1,2}\thanks{Corresponding author.}, Bineng Zhong\textsuperscript{\rm 1,2}$^*$, Shuimu Zeng\textsuperscript{\rm 3}, \\ Yuanliang Xue\textsuperscript{\rm 4}, Ning Li\textsuperscript{\rm 1,2}, Shuxiang Song\textsuperscript{\rm 1,2}\\
\textsuperscript{\rm 1}Key Laboratory of Education Blockchain and Intelligent Technology, Ministry of Education\\ Guangxi Normal University, Guilin 541004, China\\
\textsuperscript{\rm 2}University Engineering Research Center of Educational Intelligent Technology\\ Guangxi Normal University, Guilin 541004, China\\
\textsuperscript{\rm 3}University of Southampton, Southampton,  SO17 1BJ, United Kingdom\\
\textsuperscript{\rm 4}Xi’an Research Institute of High Technology, Xi’an 710025, China\\
{\tt\small yaozongzheng@stu.gxnu.edu.cn, qhliang@gxnu.edu.cn, bnzhong@gxnu.edu.cn}
}

\begin{document}
\maketitle
\begin{abstract}

Learning robust contextual knowledge from unlabeled videos is essential for advancing self-supervised tracking. However, conventional self-supervised trackers lack effective context modeling, while existing context association methods based on non-semantic queries struggle to adapt to unlabeled tracking scenarios, making it difficult to learn reliable contextual cues. In this work, we propose a novel self-supervised tracking framework, named \textbf{\tracker}, which introduces a dual-modal context association mechanism that jointly leverages fine-grained semantic prompts and contextual noise to drive the model toward learning robust tracking representations. Adherent to the easy-to-hard learning principle, our contextual association mechanism operates based on two stages. During early training, instance patch tokens (prompts) are assigned to both forward and backward tracking branches to facilitate the acquisition of tracking knowledge. As training progresses, contextual noise is gradually injected into the model to perturb feature, encouraging the tracker to learn robust tracking representations in a more complex feature space. Thus, this novel contextual association mechanism enables our self-supervised model to learn high-quality tracking representations from unlabeled videos, while being applied exclusively during training to preserve efficient inference. Extensive experiments demonstrate the superiority of our method.





\end{abstract}    
\section{Introduction}
\label{sec:intro}

   \begin{figure}[t]
      \centering
      \includegraphics[width=0.8\columnwidth]{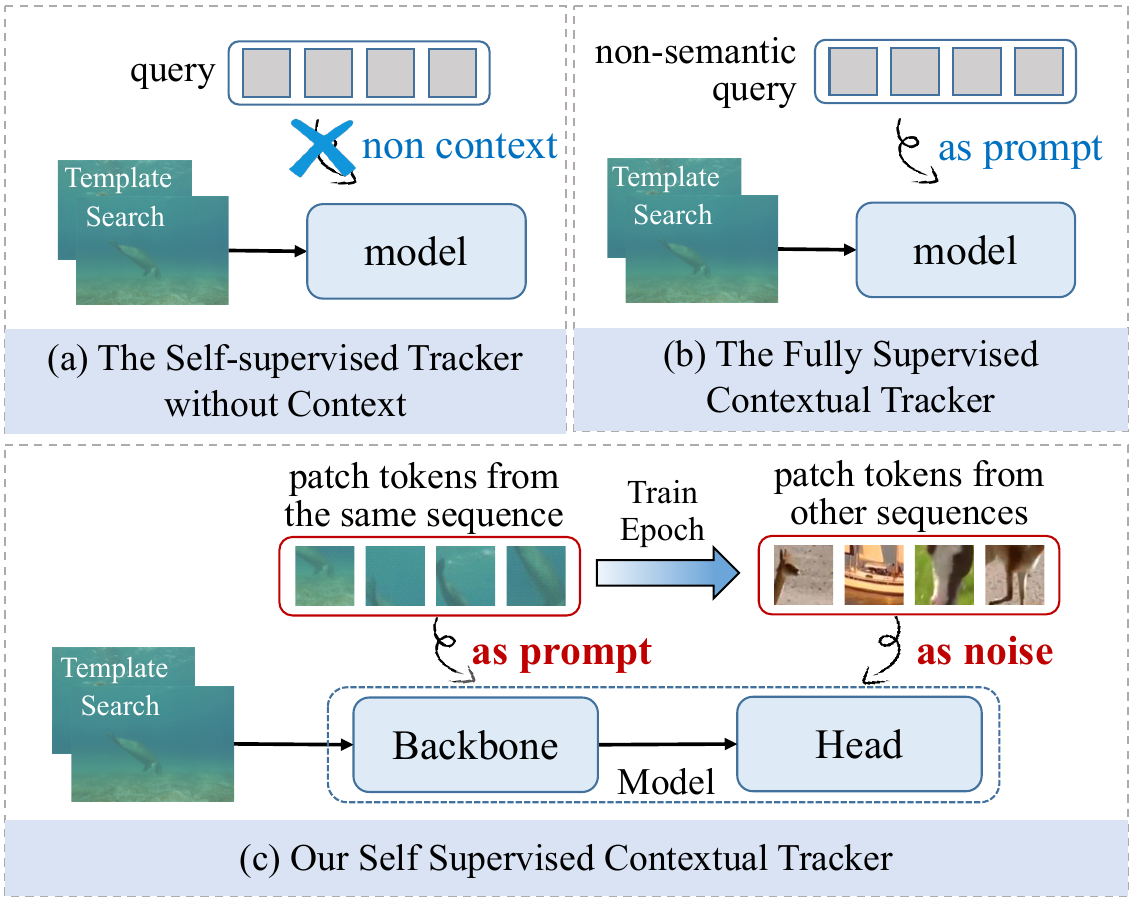}
       \caption{Comparison of different tracking methods. (a) The vanilla self-supervised tracker \cite{TADS} without contextual modeling. (b) The fully supervised contextual tracker \cite{SPMTrack,odtrack} using randomly initialized non-semantic queries. (c) Our self-supervised contextual tracker guided by prompt and noise tokens.
       }
       \label{fig:motivation}
    \end{figure}

Visual object tracking aims to continuously locate and track a specific target object throughout a video sequence. In recent years, this task has made significant strides, largely driven by the availability of large-scale annotated tracking datasets and advancements in deep learning techniques. Benchmark datasets such as LaSOT \cite{lasot}, TrackingNet \cite{trackingnet}, and GOT10k \cite{got10k} have played a crucial role in training effective tracking models, substantially contributing to performance improvements across various challenging scenarios. However, due to their reliance on meticulously human-annotated video sequences, traditional fully supervised tracking models \cite{ostrack,SPMTrack} face limitations in scalability and generalization, which hinders their broader deployment. To address this issue, self-supervised tracking has been proposed and has since garnered considerable attention. Typically, it seeks to leverage large volumes of unlabeled video frames along with a minimal amount of labeled data to alleviate the labor-intensive annotation process.

With the introduction of cycle consistency \cite{cycle-consistency} and contrastive learning \cite{InfoNCE} into the self-supervised tracking community, mainstream self-supervised tracking algorithms \cite{SDCT,cycleSiam} have evolved into widely adopted cycle consistency tracking frameworks. When dealing with a video sequence, consistency constraint aims to ensure coherent predictions for the same target object across different time steps.
For example, recent study \cite{SSTrack} based on the principle of cycle consistency decouples forward and backward tracking into global and local tracking, respectively, to handle unlabeled and labeled video frames separately, thus improving self-supervised tracking performance.

However, as illustrated in Fig.\ref{fig:motivation}(a), these self-supervised trackers lack explicit context modeling, making it difficult for them to learn context-aware cues to guide the tracking process. Meanwhile, as shown in Fig.\ref{fig:motivation}(b), directly transferring the conventional query association paradigm to the self-supervised tracking is suboptimal. We observe that the randomly initialized, non-semantic queries fail to extract sufficient and reliable contextual cues from unlabeled video frames (see Sec.\ref{sec:study}). These limitations tend to undermine the generalization ability of self-supervised trackers, particularly in long-term and complex scenarios.



To address the above challenges, we propose a novel self-supervised tracking framework, termed \textbf{\tracker}, which introduces a dual-mode contextual association mechanism designed to efficiently learn tracking representations from unlabeled videos. 
Inspired by the \textit{easy-to-hard} learning philosophy, we introduce different types of contextual information into the forward and backward tracking branches in a progressively evolving manner. The contextual information injected can originate from instance patch tokens or be randomly sampled from background tokens. Specifically, as shown in Fig.\ref{fig:motivation}(c), our dual-mode contextual association mechanism operates based on two stages. i) During early training, instance patch tokens (prompts) from video frames are assigned to both forward and backward tracking branches to facilitate the acquisition of tracking knowledge. ii) As training progresses, background tokens (noise) is gradually injected into the model to perturb feature, encouraging the tracker to learn robust tracking representations in a more complex feature space.
This novel contextual learning mechanism is applied exclusively during training and encourages the self-supervised tracker to extract valuable target cues from diverse feature spaces. As a result, it enables the model to learn high-quality tracking representations from unlabeled videos, while maintaining efficient inference during testing.
Extensive experiments on multiple tracking benchmarks demonstrate that our method achieves new \textit{SOTA} tracking performance with limited annotations. 
The main contributions are as follows.
    \begin{itemize}
    \item We make the first attempt to propose a novel context-associated self-supervised tracking framework, termed {\tracker}, which aims to further bridge the performance gap between self- and fully supervised tracking.

    \item We propose an effective dual-mode contextual association mechanism that injects diverse contextual cues into forward and backward tracking in a progressively evolving manner. This method facilitates the effective learning of robust representations from unlabeled videos.
    
    \item Our method achieves new \textit{SOTA} results on eight tracking benchmarks, including GOT10K, LaSOT, LaSOT$_{\rm{ext}}$, TrackingNet, VOT2020, TNL2K, UAV123 and OTB100.
    
    \end{itemize}

\section{Related Work}
\label{sec:works}

\subsection{Context-aware Tracking Methods}
Benefiting from the widespread availability of large-scale annotated datasets \cite{lasot,trackingnet,got10k}, fully-supervised tracking algorithms \cite{UMODTrack,SiamPIN,MMTrack} have made significant progress. Recent studies demonstrate that improving temporal contextual dependencies can substantially improve tracking performance. Most of these approaches adopt auto-regressive modeling or temporal aggregation strategies to better capture appearance and motion cues within video sequences. For instance, ARTrack \cite{ARTrack} and ARTrackV2 \cite{Artrackv2} leverage auto-regressive mechanism to jointly model appearance and trajectory information, exhibiting strong robustness in various challenging scenarios. ODTrack \cite{odtrack} introduces a video-level tracking framework with online temporal tokens, paving the way for new advancements in the field. AQATrack \cite{AQATrack} further improves performance by incorporating an attention-guided auto-regressive query mechanism.
DreamTrack \cite{DreamTrack} enhances the generalization of the tracker across diverse scenarios by predicting the future states of the target.

However, existing approaches heavily rely on non-semantic queries with random initialization to associate objects across frames, which inherently limits their representational capacity and hinders the extraction of reliable contextual cues from unlabeled videos. In this work, we attempt to exploit semantically meaningful tokens derived from video frames to guide and associate target instance under unlabeled settings, thereby improving the robustness and generalization of self-supervised tracker.

   \begin{figure*}
      \centering
      \includegraphics[width=0.8\linewidth]{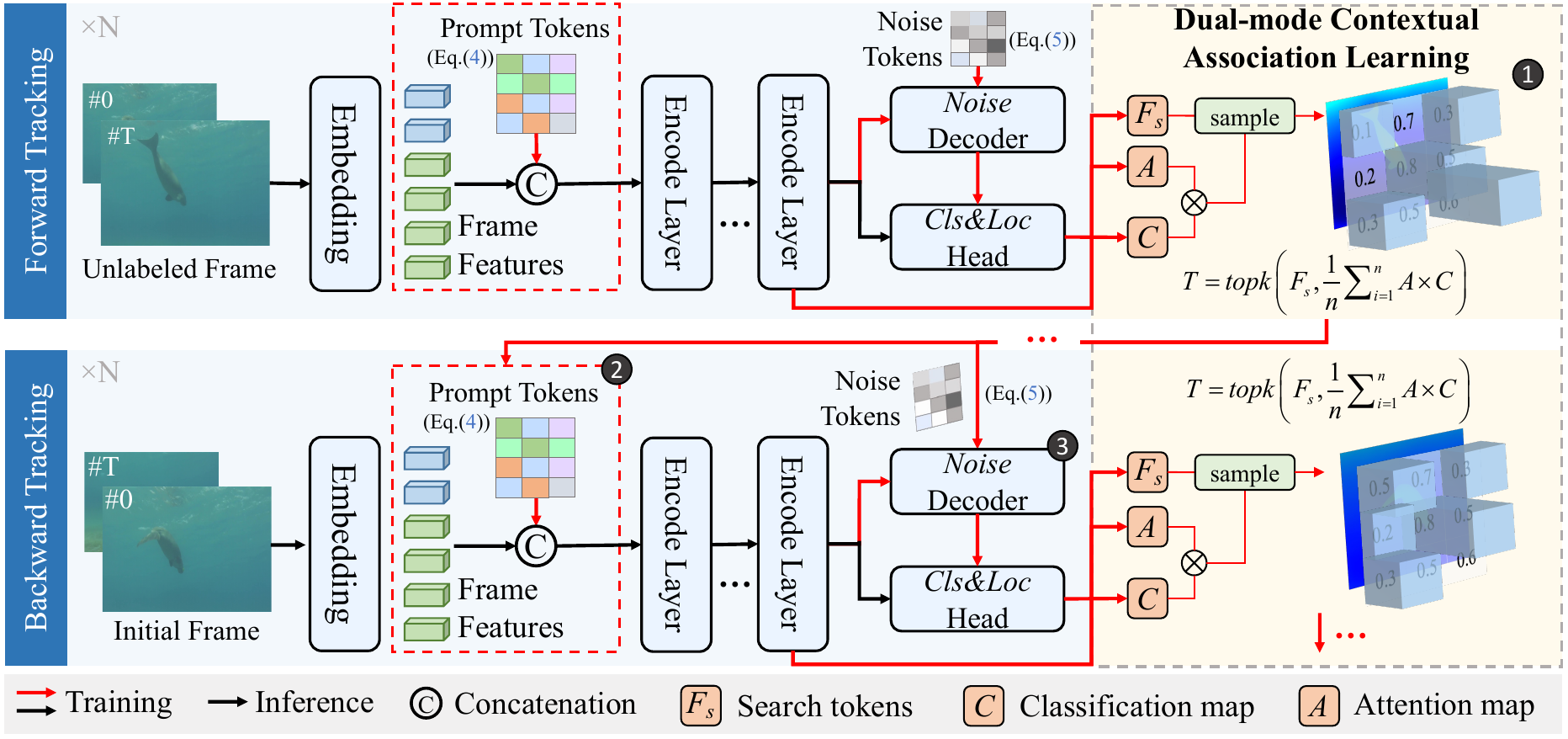}
       \caption{Overview of {\tracker} pipeline. Our dual-mode contextual association mechanism introduces distinct signals to forward and backward tracking to modulate their learning difficulties.
       During the early training phase, we utilize target state distributions derived from attention and classification response maps to sample target tokens from the current frame as prompts, thereby improving the learning efficiency of the self-supervised tracker. In contrast, in the later training phase, we inject contextual noise into both tracking branches to perturb the feature space and elevate the learning difficulty, which further encourages the model to acquire more robust tracking representations.
       }
       \label{fig:framework}
    \end{figure*}

\subsection{Self-Supervised Tracking Methods}
Self-supervised tracking has been attracting increasing attention from the research community, owing to its efficiency in label usage and scalability with large-scale data. 
CycleSiam \cite{cycleSiam} builds upon the idea of cycle consistency by integrating a region proposal network into a siamese architecture for self-supervised tracking. self-SDCT \cite{SDCT} proposes a multi-cycle consistency loss combined with a low-similarity dropout operation, encouraging the model to maintain stable representations.
S\(^2\)SiamFC \cite{s2siamfc} constructs positive pairs by sampling a local patch and its enlarged counterpart from the same image, while employing an adversarial masking strategy to improve target matching capability.
TADS \cite{TADS} enhances the diversity of training image pairs through a series of data augmentation operations, enabling self-supervised tracking in a plug-and-play manner when integrated into existing frameworks such as SiamRPN++ \cite{SiamRPN++} and TransT \cite{transt}.
Recently, SSTrack \cite{SSTrack} introduces a decoupled spatio-temporal consistency training framework and employs an instance-level contrastive loss to learn fine-grained correspondences across multiple views, leading to a significant improvement in self-supervised tracking performance.

However, in the absence of effective annotations, these methods are susceptible to noise during training and constrain the model’s capacity to adapt to challenging tracking scenarios. This oversight hinders the model's ability to fully exploit unlabeled videos for learning reliable target correspondences. 
To address this challenge, we introduce effective dual-mode contextual association learning into self-supervised tracking. By injecting diverse types of contextual tokens, our method dynamically modulates the learning difficulty of forward and backward tracking, facilitating more efficient learning of discriminative tracking features.

\section{Approach}
In this section, we will provide a detailed introduction to
the proposed {\tracker}. Before presenting the proposed framework, we first conduct a detailed analysis of the inherent challenges in self-supervised tracking. Then, we provide a detailed introduction to the dual-mode contextual association mechanism. Finally, we describe the prediction head network and the optimization objectives.

\subsection{Revisting Self-Supervised Tracking}

Self-supervised tracking algorithms aim to leverage sparse annotation from initial video frame $\mathcal{D}^{l}=\{ (z^{l}_{0}, y^{l}_{0}) \}$ to effectively exploit the semantic content of unlabeled video data $\mathcal{D}^{u}=\{ x^{u}_{i} \}$. They formulate visual tracking as a cross-temporal matching problem and learn a temporal correlation map using deep models $f_{\theta}(\cdot)$ based on siamese networks \cite{SiamRPN++} or transformer architectures \cite{ostrack}. A typical self-supervised tracking model consists of a forward tracking branch and a backward tracking branch, trained jointly via a consistency constraint. Formally, these methods \cite{s2siamfc,SSTrack} perform forward tracking $\mathcal{E}$ to process unlabeled frames, and subsequently use the predictions from these frames during backward tracking $\phi$ to infer the target location in the initial frame over a larger search region, thus completing a training epoch:
   \begin{equation}
     \begin{split}
     \mathcal{B}^u_{1:t} &= \mathcal{E} \left(z^{l}_{0}, x^u_{1:t} \right), \\
        \mathcal{B}^l_{0} &= \phi \left( \sigma \left(\mathcal{B}^u_{1:t}, x^u_{1:t} \right), x^{l}_{0} \right),
      \end{split}
     \label{eq:model}
   \end{equation}
where $\sigma(\cdot, \cdot)$ denotes a cropping operation performed on unlabeled search frames, guided by forward tracking results, to generate new reference frames for backward tracking.

However, this simple modeling method implies an \textit{intrinsic} restrictions in the design of self-supervised trackers. From an information-theoretic perspective, the goal of the self-supervised tracking task is to maximize the mutual information $I$ between the input video sequence (i.e., $\mathcal{D} = \{ \mathcal{D}^l, \mathcal{D}^u \}$) and the target state $y_0^l$, i.e.,
    \begin{equation}
        \max_{\theta} I \left( f_{\theta} \left(\mathcal{D} \right); y_0^l \right).
        \label{eq:max_info}
    \end{equation}
However, this self-supervised contextual learning process is often constrained by the limited labeled subset data $\mathcal{D}^l$ and the non-semantic token association manner. This limitation tends to yield unreliable contextual information, where the mutual information $I(\mathcal{F}; y_0^l)$ between the learned feature representation $\mathcal{F}$ and the target signal $y_0^l$ is restricted. Consequently, the model fails to fully exploit the rich target information embedded in the unlabeled dataset $\mathcal{D}^u$, thereby limiting its generalization ability in complex scenarios.


To address the aforementioned challenge, this work proposes a novel self-supervised tracking framework, named {\tracker}, which aims to learn robust tracking representations more effectively from unlabeled videos. Specifically, we design a flexible dual-mode contextual association mechanism within the self-supervised tracking framework to adaptively regulate the learning difficulty during training. By progressively injecting diverse contextual signals into the forward and backward tracking branches, we explicitly guide the model to first learn from simpler scenarios and then gradually adapt to more complex ones. The next section provides a detailed description of this mechanism.

\subsection{Dual-mode Contextual Association Mechanism}

Previous approaches typically focus on techniques such as contrastive learning to achieve effective self-supervised tracking. However, these trackers are inefficient in handling complex and diverse unlabeled scenarios due to the lack of dynamic modulation in the network learning strategy. To better capture the latent semantic content in unlabeled videos, it is crucial to construct a flexible flow of contextual information. As illustrated in Fig.\ref{fig:framework}, we propose a Dual-mode Contextual Association (DCA) mechanism that adaptively allocates contextual information to the forward and backward tracking branches, thereby modulating the learning difficulty throughout training. 
Specifically, we regard not only the \textit{target tokens} but also the \textit{background tokens} as forms of contextual information. 
target (prompt) tokens define what to focus on, while background (noise) tokens regulates what to suppress. Both offer complementary mechanisms for context modeling.
Following a temporal scheduling principle, our DCA determines whether to propagate prompt tokens or noise tokens to each branch at different training stages.

\begin{algorithm}[tb]
\caption{The training process of {\tracker}}
\label{alg:algorithm}
\textbf{Input}: $\mathcal{D}^{l}=\{ (z^{l}_{0}, y^{l}_{0}) \}$; $\mathcal{D}^{u}=\{ x^{u}_{i} \}$ \\
\textbf{Output}: Tracking results $B^l_i$
\begin{algorithmic}[1] 
\STATE // \textit{Forward tracking}
\STATE T = None
\FOR{i = 1 to L}
\STATE $\mathcal{F}_{x}, \mathcal{A}, \mathcal{C}, B^u_{1:i} = \mathcal{E} \left(z^{l}_{0}, x^u_{1:i}, T \right)$
\IF{epoch $\le$ K}
\STATE Extract tokens $T = \text{topk} \left( \mathcal{F}_{x}, \frac{1}{n} \sum_{j=1}^{n}{\mathcal{A} \times \mathcal{C}} \right)$
\ELSE
\STATE Extract tokens $T = \text{random} \left( \mathcal{F}_{x} \right)$
\ENDIF
\ENDFOR
\STATE // \textit{Backward tracking}
\STATE T = None
\FOR{i = 1 to m}
\STATE $\mathcal{F}_{x}, \mathcal{A}, \mathcal{C}, B^l_0 = \phi \left( \sigma \left(\mathcal{B}^u_{1:i}, x^u_{1:i} \right), x^{l}_{0}, T \right)$
\IF{epoch $\le$ K}
\STATE Extract tokens $T = \text{topk} \left( \mathcal{F}_{x}, \frac{1}{n} \sum_{j=1}^{n}{\mathcal{A} \times \mathcal{C}} \right)$
\ELSE
\STATE Extract tokens $T = \text{random} \left( \mathcal{F}_{x} \right)$
\ENDIF
\ENDFOR
\STATE Calculate loss using Eq.\ref{eq:loss} and update parameters.
\end{algorithmic}
\end{algorithm}

\textbf{Contextual Prompt for Accelerated Learning.}
We adopt a visual transformer (i.e., ViT \cite{vit}) as the backbone network for feature extraction and fusion. Theoretically, the absence of sufficient supervision makes it highly challenging to accurately localize target instances in unlabeled frames solely based on the network’s decision output, such as the bounding box with the highest classification score. However, recent studies \cite{attention,DETR} have revealed that attention maps inherently possess the potential to indicate the spatial location of target objects within an image. This insight enables the possibility of leveraging powerful attention maps as auxiliary signals for target localization and tracking.

First, in the {\tracker} framework, our forward and backward tracking processes can be described as follows:
   \begin{equation}
     \begin{split}
     \mathcal{F}_{x}, \mathcal{A}, \mathcal{C}, B^u_{1:i} &= \mathcal{E} \left(z^{l}_{0}, x^u_{1:i}, T \right) \\
     &= \{Attn \left( z^{l}_{0}, x^u_{1:i}, T \right) \}^{K}_{i=1}, K=12,\\
     \mathcal{F}_{x}, \mathcal{A}, \mathcal{C}, B^l_0 &= \phi \left( z^u_{1:t}, x^{l}_{0}, T \right) \\
     &= \{Attn \left( z^u_{1:t}, x^{l}_{0}, T \right) \}^{K}_{i=1}, K=12,\\
      \end{split}
     \label{eq:vit}
   \end{equation}
where $\mathcal{F}_{x}$ denotes the output features of the search frame. $\mathcal{A}$ denotes the cross-attention weight matrix between the reference frame and the search frame, which captures the state distribution of the tracked object within the current search frame. $\mathcal{C}$ represents the spatial localization of the target object in the search frame. $\mathcal{B}$ is the predicted bounding box.

Then, the DCA mechanism primarily serves to extract the contextual information of each video frame and propagate it to the subsequent frame. In the early training stage, the DCA is designed to extract semantically meaningful contextual prompts for cross-frame association, thereby accelerating the learning efficiency of the self-supervised tracking model. Specifically, DCA first caches a spatial attention map $\mathcal{A}$ from the last transformer layer of the backbone and performs a cross-correlation operation with the classification response map $\mathcal{C}$ to reinforce the semantic content of each token within the video frames. Furthermore, based on the weight ranking results of the learned cross-correlation response map, the top K tokens are sampled from the search features $\mathcal{F}_{s}$ as instance tokens. Formally, the sampling process of instance tokens $T$ is defined as:
\begin{equation}
    T = \text{topk} \left( \mathcal{F}_{x}, \frac{1}{n} \sum_{j=1}^{n}{\mathcal{A} \times \mathcal{C}} \right),
    \label{eq:prompt}
\end{equation}
where $n$ represents the number of attention heads.
The cross-correlation $\mathcal{A} \times \mathcal{C}$ acts as a token-level assessment function that determines whether each search token corresponds to the target object or background content.
It is worth noting that, before performing the cross-correlation, the multi-head attention matrix $\mathcal{A}$ is averaged across the attention head dimension to ensure dimensional alignment.

Once high-confidence target tokens are sampled, they are fed into the forward or backward tracking branch of the next frame as contextual prompts to guide the self-supervised training.
This specific mechanism is designed to offer richer cues before the model has learned reliable tracking patterns, facilitating stable optimization in the early training stage.


\textbf{Contextual Noise for Feature Perturbation}.
Unlike the early training stage, where target prompts are employed to improve the efficiency of {\tracker} in learning tracking knowledge, background tokens (contextual noise) are injected during the \textit{later training} phase to further improve the model's robustness by perturbing the feature space. 
Specifically, instead of carefully evaluating the quality of each visual token, we randomly sample background tokens from each frame as noise tokens, encouraging the tracker to learn robust representations in complex feature distributions. Formally, we use the formula to describe:
\begin{equation}
    T = \text{random} \left( \mathcal{F}_{x} \right),
    \label{eq:noise}
\end{equation}

Furthermore, we design a noise decoder consisting of a transformer layer followed by classification and regression heads. The search features are perturbed through cross-attention with background tokens sampled from other sequence, after which the perturbed features are fed into the prediction heads for classification and regression.

\textit{Why does this feature perturbation strategy work?}
We attribute its effectiveness to two key reasons. First, after the early training stage, the tracker has already acquired a certain level of instance tracking knowledge, making it less susceptible to optimization drift caused by noisy tokens. Second, injecting contextual noise perturbs the semantic embedding space of the original search features, effectively simulating a more challenging tracking environment. This encourages {\tracker} to learn more robust target localization capabilities. 
Based on these observations, we argue that our DCA holds promise in adaptively modulating training difficulty, thus mitigating the bottleneck that hinders self-supervised tracker from efficiently learning robust tracking representations.
Importantly, the DCA is utilized only during training and is omitted at inference time.

\subsection{Head and Objective Loss}
Similar to most transformer-based trackers \cite{ostrack}, we employ stacked convolutional layers to construct the classification and regression heads. For the loss functions, we adopt focal loss $\mathcal{L}_{cls}$ \cite{focalloss} for classification, and a combination of GIoU loss $\mathcal{L}_{GIoU}$ \cite{giou} and $\mathcal{L}_1$ loss for regression. The overall loss function of {\tracker} is defined as follows:
   \begin{equation}
      \mathcal{L} = \mathcal{L}_{cls} + \lambda_{1}\mathcal{L}_{1} + \lambda_{2}\mathcal{L}_{GIoU},
     \label{eq:loss}
   \end{equation}
where $\lambda_{1}$ and $\lambda_{2}$ are the regularization parameters.
Thus, the training process of {\tracker} is shown in Algorithm \ref{alg:algorithm}.

\section{Experiments}


\subsection{Implementation Details}

The model is trained end-to-end on four training datasets: LaSOT \cite{lasot}, GOT10k \cite{got10k}, TrackingNet \cite{trackingnet}, and COCO \cite{coco}, using the AdamW optimizer to minimize the network losses. We use the same number and size of input video frames as in \cite{SSTrack}. We adopt the ViT-Base \cite{vit} as the backbone and initialize its parameters with DropMAE \cite{DropMAE} pre-trained weights. The initial learning rate is set to \(2.5 \times 10^{-5}\) for the backbone and \(2.5 \times 10^{-4}\) for the remaining components, with a weight decay of \(10^{-4}\) applied throughout training.
The training is conducted on a server equipped with two A100 GPUs, using a batch size of 8. The model is trained for 150 epochs, where $10,000$ image pairs are randomly sampled per epoch. The learning rate is decayed by a factor of 10 after 120 epochs. 
Additionally, our {\tracker}-384 runs at 59 frames per second on an A100 GPU.

\subsection{Comparison with the SOTA}

\begin{table*}[t]
    \centering
    \caption{Comparison with state-of-the-arts on four popular benchmarks: GOT10K, LaSOT, TrackingNet, and LaSOT$_{\rm{ext}}$. Where $*$ denotes for trackers only trained on GOT10K. The best two results are shown in {\color{red}red} and {\color{blue}blue} fonts.}
    \resizebox{\textwidth}{!}{
    \begin{tabular}{l|c|c|ccc|ccc|ccc|ccc}
    \toprule 
     \multicolumn{1}{c|}{\multirow{2}{*}{Method}} & \multicolumn{1}{c|}{\multirow{2}{*}{Source}} & \multicolumn{1}{c|}{\multirow{2}{*}{Supervised}} & \multicolumn{3}{c|}{GOT10K$^*$} & \multicolumn{3}{c|}{LaSOT} & \multicolumn{3}{c|}{TrackingNet} & \multicolumn{3}{c}{LaSOT$_{\rm{ext}}$} \\
     \cline{4-15}
      & & & AO & SR${_{0.5}}$ & SR${_{0.75}}$ & AUC & P${_{\rm{Norm}}}$ & P & AUC & P${_{\rm{Norm}}}$ & P & AUC & P${_{\rm{Norm}}}$ & P \\
      \midrule
      TransT \cite{transt} & CVPR21 & \checkmark & 67.1 & 76.8 & 60.9 & 64.9 & 73.8 & 69.0 & 81.4 & 86.7 & 80.3 & - & - & - \\
      Stark \cite{stark} & ICCV21 & \checkmark & 68.8 & 78.1 & 64.1 & 67.1 & 77.0 & - & 82.0 & 86.9 & - & - & - & - \\
      Mixformer \cite{mixformer} & CVPR22 & \checkmark & 70.7 & 80.0 & 67.8 & 69.2 & 78.7 & 74.7 & 83.1 & 88.1 & 81.6 & - & - & - \\
      OSTrack \cite{ostrack} & ECCV22 & \checkmark & 73.7 & 83.2 & 70.8 & 71.1 & 81.1 & 77.6 & 83.9 & 88.5 & 83.2 & 50.5 & 61.3 & 57.6 \\
      SeqTrack \cite{seqtrack} & CVPR23 & \checkmark & 74.5 & 84.3 & 71.4 & 71.5 & 81.1 & 77.8 & 83.9 & 88.8 & 83.6 & 50.5 & 61.6 & 57.5 \\
      ARTrack \cite{ARTrack} & CVPR23 & \checkmark & 75.5 & 84.3 & 74.3 & 72.6 & 81.7 & 79.1 & 85.1 & 89.1 & 84.8 & 51.9 & 62.0 & 58.5 \\
      ODTrack \cite{odtrack} & AAAI24 & \checkmark & 77.0 & 87.9 & 75.1 & 73.2 & 83.2 & 80.6 & 85.1 & 90.1 & 84.9 & 52.4 & 63.9 & 60.1 \\
      HIPTrack \cite{HIPTrack} & CVPR24 & \checkmark & 77.4 & 88.0 & 74.5 & 72.7 & 82.9 & 79.5 & 84.5 & 89.1 & 83.8 & - & - & - \\
      AQATrack \cite{AQATrack} & CVPR24 & \checkmark & 76.0 & 85.2 & 74.9 & 72.7 & 82.9 & 80.2 & 84.8 & 89.3 & 84.3 & 52.7 & 64.2 & 60.8 \\
      ARTrackV2 \cite{Artrackv2} & CVPR24 & \checkmark & 77.5 & 86.0 & 75.5 & 73.0 & 82.0 & 79.6 & 85.7 & 89.8 & 85.5 & 52.9 & 63.4 & 59.1 \\
      LMTrack \cite{LMTrack} & AAAI25 & \checkmark & 76.3 & 87.1 & 73.9 & 69.8 & 79.2 & 76.3 & 84.2 & 89.0 & 82.8 & 49.0 & 59.6 & 55.8 \\
      MCITrack \cite{MCITrack} & AAAI25 & \checkmark & 77.9 & 88.2 & 76.8 & 75.3 & 85.6 & 83.3 & 86.3 & 90.9 & 86.1 & 54.6 & 65.7 & 62.1 \\
      LoRAT-B378 \cite{lorat} & ECCV24 & \checkmark & 73.7 & 82.6 & 72.9 & 72.9 & 81.9 & 79.1 & 84.2 & 88.4 & 83.0 & 53.1 & 64.8 & 60.6 \\
      ARPTrack \cite{ARPTrack} & CVPR25 & \checkmark & 77.7 & 87.3 & 74.3 & 72.6 & 81.4 & 78.5 & 85.5 & 90.0 & 85.3 & 52.0 & 62.9 & 58.7 \\
      DreamTrack \cite{DreamTrack} & CVPR25 & \checkmark & 77.5 & 87.1 & 74.2 & 73.8 & 83.4 & 80.6 & 85.8 & 90.0 & 85.3 & 53.1 & 64.1 & 59.8 \\
      \midrule
      LUDT \cite{LUDT} & IJCV21 & \ding{55} & - & - & - & {26.2} & - & {23.4} & 56.3 & 63.3 & 49.5 & - & - & - \\
      USOT \cite{USOT} & ICCV21 & \ding{55} & - & - & - & {33.7} & - & {32.5} & 59.9 & 68.2 & 55.1 & - & - & - \\
      ULAST \cite{ULAST} & CVPR22 & \ding{55} & - & - & - & {47.1} & - & {45.1} & 65.4 & 73.2 & 59.2 & - & - & - \\
      Diff-Tracker \cite{DiffTracker} & ECCV24 & \ding{55} & - & - & - & {48.6} & - & {47.2} & 67.5 & 75.1 & 61.4 & - & - & - \\
      TADS \cite{TADS} & TNNLS23 & \textit{Init.BBox} & {46.7} & {56.5} & {21.1} & {45.5} & {54.2} & {44.8} & {65.6} & {73.4} & {60.6} & - & - & - \\
      SSTrack-256 \cite{SSTrack} & AAAI25 & \textit{Init.BBox} & 67.1 & 76.6 & 59.1 & 64.8 & 75.2 & 69.7 & 80.1 & 86.7 & 78.9 & 46.2 & 57.8 & 52.1 \\
      SSTrack-384 \cite{SSTrack} & AAAI25 & \textit{Init.BBox} & \color{blue}72.4 & \color{red}83.6 & \color{blue}66.2 & \color{blue}65.9 & \color{blue}76.4 & \color{blue}70.7 & \color{blue}80.4 & 86.3 & \color{blue}77.9 & \color{blue}48.5 & \color{blue}60.9 & \color{blue}54.5 \\
      \textbf{{\tracker}-256} & Ours & \textit{Init.BBox} & {69.0} & {79.1} & {62.7} & \color{blue}{65.9} & {76.3} & {70.2} & \color{blue}{80.4} & \color{blue}{86.7} & {77.8} & {47.0} & {58.8} & {53.1} \\
      \textbf{{\tracker}-384} & Ours & \textit{Init.BBox} & \color{red}{72.7} & \color{blue}{83.4} & \color{red}{69.9} & \color{red}{67.1} & \color{red}{77.3} & \color{red}{71.9} & \color{red}{81.8} & \color{red}{87.4} & \color{red}{79.7} & \color{red}{49.1} & \color{red}{61.2} & \color{red}{55.1} \\
    \bottomrule
    \end{tabular} }
    \label{tab:results}
\end{table*}

\begin{table*}[t]
\centering
\caption{Comparison with sota methods on multiple benchmarks in AUC score. The best two results are shown in {\color{red}red} and {\color{blue}blue} fonts.
}
\resizebox{\textwidth}{!}{
\begin{tabular}{l|ccccccc|ccccc}
\toprule
\multicolumn{1}{c|}{\multirow{3}{*}{Datesets}} & \multicolumn{7}{c|}{Fully Supervised Tracking} & \multicolumn{5}{c}{Self Supervised Tracking} \\
\cline{2-13}
& \begin{tabular}[c]{@{}c@{}}DiMP\\ \cite{DiMP50}\end{tabular} & \begin{tabular}[c]{@{}c@{}}TransT\\ \cite{transt}\end{tabular} & \begin{tabular}[c]{@{}c@{}}TrDiMP\\ \cite{trdimp}\end{tabular} & \begin{tabular}[c]{@{}c@{}}Mixformer\\ \cite{mixformer}\end{tabular} & \begin{tabular}[c]{@{}c@{}}HIPTrack\\ \cite{HIPTrack}\end{tabular} & \begin{tabular}[c]{@{}c@{}}LoRAT-L378\\ \cite{lorat}\end{tabular} & \begin{tabular}[c]{@{}c@{}}DreamTrack\\ \cite{DreamTrack}\end{tabular} & \begin{tabular}[c]{@{}c@{}}self-SDCT\\ \cite{SDCT}\end{tabular} & \begin{tabular}[c]{@{}c@{}}TADS\\ \cite{TADS}\end{tabular} & \begin{tabular}[c]{@{}c@{}}SSTrack-256\\ \cite{SSTrack}\end{tabular} & \begin{tabular}[c]{@{}c@{}}\textbf{{\tracker}-256}\\ Ours\end{tabular} & \begin{tabular}[c]{@{}c@{}}\textbf{{\tracker}-384}\\ Ours\end{tabular} \\
\midrule
TNL2K(AUC) & 44.7 & 50.7 & - & - & - & 62.3 & 61.2 & - & - & 52.1 & \color{blue}{53.4} & \color{red}{55.3} \\
OTB100(AUC) & 68.4 & 69.4 & 67.5 & 70.0 & 71.0 & 72.0 & 72.0 & 63.8 & 65.3 & 67.9 & \color{blue}{68.8} & \color{red}{71.2} \\
UAV123(AUC) & 65.3 & 69.1 & 67.5 & 70.4 & 70.5 & 72.5 & 72.5 & 50.1 & 55.2 & 65.5 & \color{blue}{65.9} & \color{red}{66.4} \\
\bottomrule
\end{tabular} }
\label{tab:tnl2k}
\end{table*}

\begin{table}[t]
\centering
\caption{State-of-the-art comparison on VOT2020 benchmark. The best two results are shown in {\color{red}red} and {\color{blue}blue} fonts.}
\resizebox{\linewidth}{!}{
\begin{tabular}{l|ccc}
\toprule
Method & EAO & Accuracy & Robustness \\
\midrule
\multicolumn{4}{l}{\multirow{1}{*}{\textit{Fully Supervised}}} \\
Ocean \cite{Ocean} & 0.430 & 0.693 & 0.754 \\
AlphaRef \cite{Alpha-Refine} & 0.482 & 0.754 & 0.777 \\
STARK \cite{stark} & 0.505 & 0.759 & 0.819 \\
SBT \cite{SBT} & 0.515 & 0.752 & 0.825 \\
Mixformer \cite{mixformer} & 0.535 & 0.761 & 0.854 \\
SeqTrack \cite{seqtrack} & 0.522 & - & - \\
ODTrack \cite{odtrack} & 0.581 & 0.764 & 0.877 \\
LMTrack \cite{LMTrack} & 0.586 & 0.753 & 0.895 \\
\midrule
\multicolumn{4}{l}{\multirow{1}{*}{\textit{Self Supervised}}} \\
SSTrack-256 \cite{SSTrack} & 0.458 & 0.664 & 0.839 \\
\textbf{\tracker-256} & \color{blue}{0.468} & \color{blue}{0.689} & \color{blue}{0.825} \\
\textbf{\tracker-384} & \color{red}{0.522} & \color{red}{0.754} & \color{red}{0.832} \\
\bottomrule
\end{tabular}}
\label{tab:vot20}
\end{table}



\textbf{GOT10K} \cite{got10k} is a visual tracking benchmark that requires evaluation under a one-shot protocol. As shown in Tab.\ref{tab:results}, we compare {\tracker}, with both self- and fully-supervised trackers. Compared to the latest self-supervised model SSTrack-256, {\tracker}-256 achieves improvements of 1.9\%, 2.5\%, and 3.6\% on AO, SR$_{0.5}$, and SR$_{0.75}$, respectively.
These results are primarily attributed to the dual-mode contextual association mechanism, which enables progressive and effective learns instance tracking knowledge from unlabeled videos.

\textbf{TrackingNet} \cite{trackingnet} is a large-scale short-term tracking benchmark comprising 511 test video sequences. As shown in Tab.\ref{tab:results}, our method achieves a new state-of-the-art result in the self-supervised tracking field. For instance, {\tracker}-384 outperforms SSTrack-384 by 1.4\%, 1.1\%, and 1.8\% in terms of AUC, P$_{\text{Norm}}$, P, respectively. This demonstrates that the introduced background tokens (contextual noise) improves the robustness of {\tracker} in complex scenarios.

\textbf{LaSOT \cite{lasot} and LaSOT$_{\rm{ext}}$ \cite{lasot-ext}} serve as fundamental benchmarks for long-term visual tracking. As shown in Tab.\ref{tab:results}, our {\tracker}-384 achieves success score improvements of 1.2\% and 0.6\% over SSTrack-384 on the LaSOT and LaSOT$_{\rm{ext}}$ datasets, respectively. Furthermore, our self-supervised tracker substantially narrows the performance gap with fully supervised methods, with the AUC score gap is reduced to 8.2\% and 5.5\% on the LaSOT and LaSOT$_{\rm{ext}}$ datasets, respectively. These results highlight that the sampled contextual prompts facilitate learning of long-term appearance variations from unlabeled videos, thus improving the robustness of {\tracker} in extended tracking scenarios.

\begin{figure}[t]
      \centering
      \includegraphics[width=0.8\linewidth]{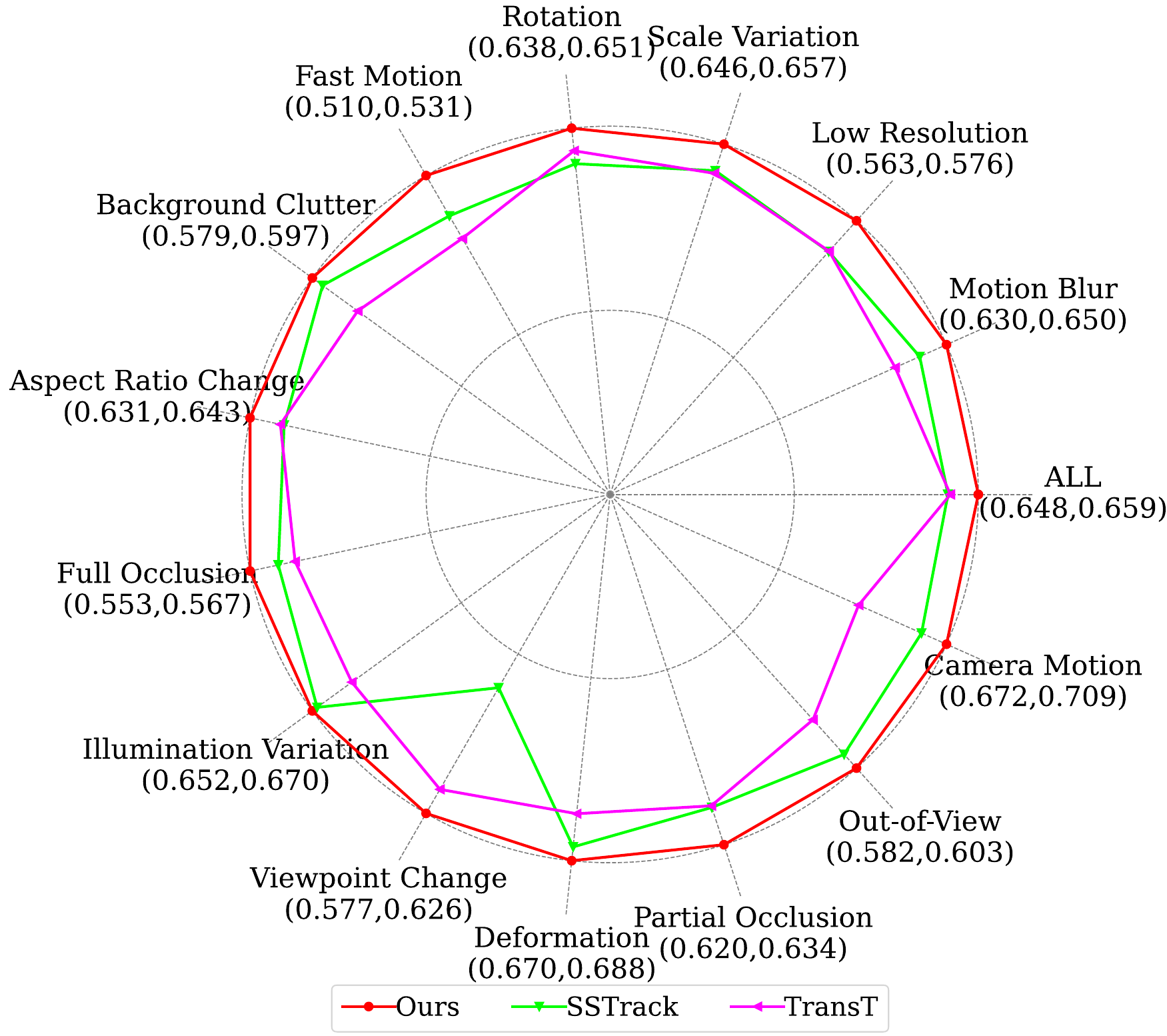}
       \caption{AUC scores of different attributes on LaSOT.
       }
       \label{fig:attr}
    \end{figure}

Furthermore, as illustrated in the Fig.\ref{fig:attr}, we analyze the attribute results of the proposed tracker across all challenges on the LaSOT benchmark, including motion blur, fast motion, occlusion, and so on. Our {\tracker} consistently outperforms both the fully supervised TransT tracker and the self-supervised SSTrack tracker on multiple challenging attributes, demonstrating the effectiveness and strong generalizability of the proposed method.

\textbf{TNL2K \cite{tnl2k} and VOT2020 \cite{VOT2020}} are large-scale benchmark datasets that encompass diverse tracking scenarios. As shown in Tab.\ref{tab:tnl2k} and Tab.\ref{tab:vot20}, our proposed self-supervised model consistently outperforms most fully supervised and self-supervised trackers. For example, {\tracker}-384 achieves AUC and EAO scores of 55.3\% and 0.522 on the TNL2K and VOT2020 datasets, respectively. These results demonstrate the effectiveness of our dual-mode contextual association learning in flexibly acquiring rich tracking representations from unlabeled video frames, thus enhancing the robustness of the model.

\textbf{OTB100 \cite{OTB2015} and UAV123 \cite{uav123}} are classical visual tracking benchmarks that present various challenges such as occlusion, distractors with similar appearance, and small targets. As shown in Tab.\ref{tab:tnl2k}, our self-supervised tracking approach achieves performance on par with most fully supervised methods. Specifically, {\tracker}-384 obtains AUC scores of 71.2\% and 66.4\% on the OTB100 and UAV123 datasets, respectively, outperforming SSTrack by 0.7\% and 0.3\%. These performance gains indicate that {\tracker}, equipped with the dual-mode contextual association mechanism, effectively leverages contextual noise to improve the generalization capability of the model.

\subsection{Ablation Study}
\label{sec:study}

To investigate the contributions of individual components, we conduct extensive experiments on the LaSOT dataset.

\textbf{Study on different association methods}.
We compare the impact of different context association strategies on the self-supervised tracking model. As shown in Tab.\ref{tab:module}, when using query vector as the cross-frame association medium (\#2), the AUC, P$_{Norm}$, and P scores drop by 1.3\%, 2.2\%, and 1.9\%, respectively. These results indicate that, compared with semantic (patch) tokens derived from real video frames, the randomly initialized non-semantic queries possess weaker representational capacity and lack sufficient and reliable contextual cues for self-supervised tracking.

\textbf{Importance of the DCA mechanism}.
Our proposed DCA comprises two core components: contextual prompts and contextual noise, where the latter is used to perturb the feature space. As shown in Tab.\ref{tab:module}, removing contextual noise from the model (\#3) leads to a 0.8\% drop in AUC score, indicating that feature perturbation is effective in enhancing the model’s ability to learn robust tracking representations. Furthermore, we examine the importance of contextual prompts to the self-supervised model. When these sampled prompts are discarded (\#4), the AUC drops by 1.8\%, suggesting that propagating contextual prompts helps reinforce the spatio-temporal consistency of target representations.
Lastly, we analyze the effect of attention maps on the sampling of instance tokens. Replacing attention-based sampling with classification confidence map (\#5) leads to a 1.1\% decrease in AUC score. This degradation is primarily due to diminished spatial awareness in the model’s ability to accurately localize the target.

\begin{table}[h]
  \centering
  \caption{Ablation studies of our tracker variants on LaSOT.}
  \begin{tabular}{c|c|ccc}
    \toprule
    \# & Method & AUC & P$_{Norm}$ & P \\
    \midrule
    1 & \textit{\tracker} & 65.9 & 76.3 & 70.2 \\
    2 & \textit{Non-semantic Query} & 64.6 & 74.1 & 68.3 \\
    \bottomrule
    3 & - \textit{Contextual Noise} & 65.1 & 75.5 & 68.9 \\
    4 & - \textit{Contextual Prompts} & 64.1 & 74.9 & 68.6 \\
    5 & - \textit{Attention Map} & 64.8 & 75.6 & 69.5 \\
    \bottomrule
  \end{tabular}
  \label{tab:module}
\end{table}


\textbf{Study on the length of the contextual token.}
As presented in the Tab.\ref{tab:prompt}, we investigate the effect of varying the length of the contextual token on self-supervised tracking performance. In particular, increasing the token length from 4 to 8 yields a 0.5\% improvement in AUC. However, further increasing the length leads to a decline in performance, indicating that selecting an appropriate and precise number of target tokens is essential during the early training stage. Excessive token sampling may introduce noise from non-target regions, thus undermining training stability.


\begin{table}[h]
  \centering
  \caption{Comparison of different contextual token lengths.}
  \begin{tabular}{c|c|ccc}
    \toprule
    \# & Token Length & AUC & P$_{Norm}$ & P \\
    \midrule
    1 & \textit{4} & 65.4 & 75.6 & 69.7 \\
    2 & \textit{8} & 65.9 & 76.3 & 70.2 \\
    3 & \textit{16} & 65.3 & 75.6 & 69.3 \\
    \bottomrule
  \end{tabular}
  \label{tab:prompt}
\end{table}

\textbf{Study on the start epoch of the feature perturbation.} 
We conduct experiments in Tab.\ref{tab:perturb} to investigate the impact of feature perturbation at different training epochs on our model. When the perturbation epoch increases from 50 to 75, the AUC score improves by 0.7\%. Further extending the perturbation epochs leads the tracking performance to gradually converge within a stable range. However, if noise tokens are injected into the model from the beginning of training, the AUC score drops by 0.7\%. We observe that selecting an appropriate perturbation epoch facilitates the learning of robust representations from unlabeled videos.

\begin{table}[h]
  \centering
  \caption{Comparison of different feature perturbation epochs.}
  \begin{tabular}{c|c|ccc}
    \toprule
    \# & Perturbation Epoch & AUC & P$_{Norm}$ & P \\
    \midrule
    1 & \textit{0} & 64.5 & 75.0 & 68.6 \\
    2 & \textit{50} & 65.2 & 75.3 & 69.3 \\
    3 & \textit{75} & 65.9 & 76.3 & 70.2 \\
    4 & \textit{100} & 65.6 & 75.7 & 69.8 \\
    \bottomrule
  \end{tabular}
  \label{tab:perturb}
\end{table}

\subsection{Qualitative Analysis}

\begin{figure}[t]
      \centering
      \includegraphics[width=0.75\linewidth]{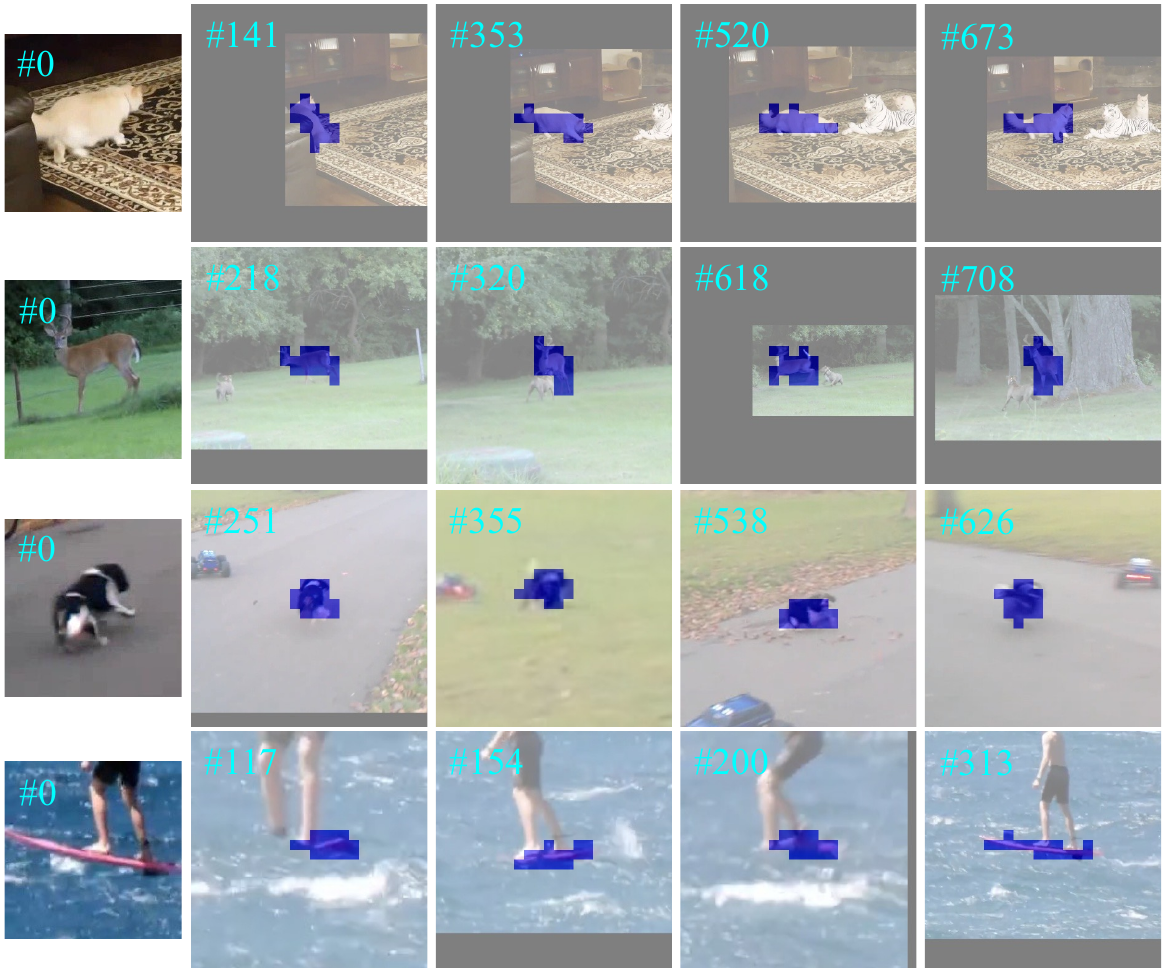}
       \caption{Visualization of the contextual prompts (object tokens) sampled from the LaSOT benchmark.
       }
       \label{fig:visual}
    \end{figure}

To intuitively validate whether the contextual prompts sampled genuinely originate from the target instance, we conduct a comprehensive qualitative analysis in Fig.\ref{fig:visual}. Specifically, in the second sequence, when a dog approaches the target (a deer), the sampled target tokens effectively avoid including the dog’s body, indicating that our contextual prompt sampling method possesses strong target-awareness capability. Moreover, this modeling approach facilitates reliable cross-frame associations, enabling {\tracker} to perform robustly in various complex tracking scenarios.

\begin{figure}[t]
      \centering
      \includegraphics[width=0.75\linewidth]{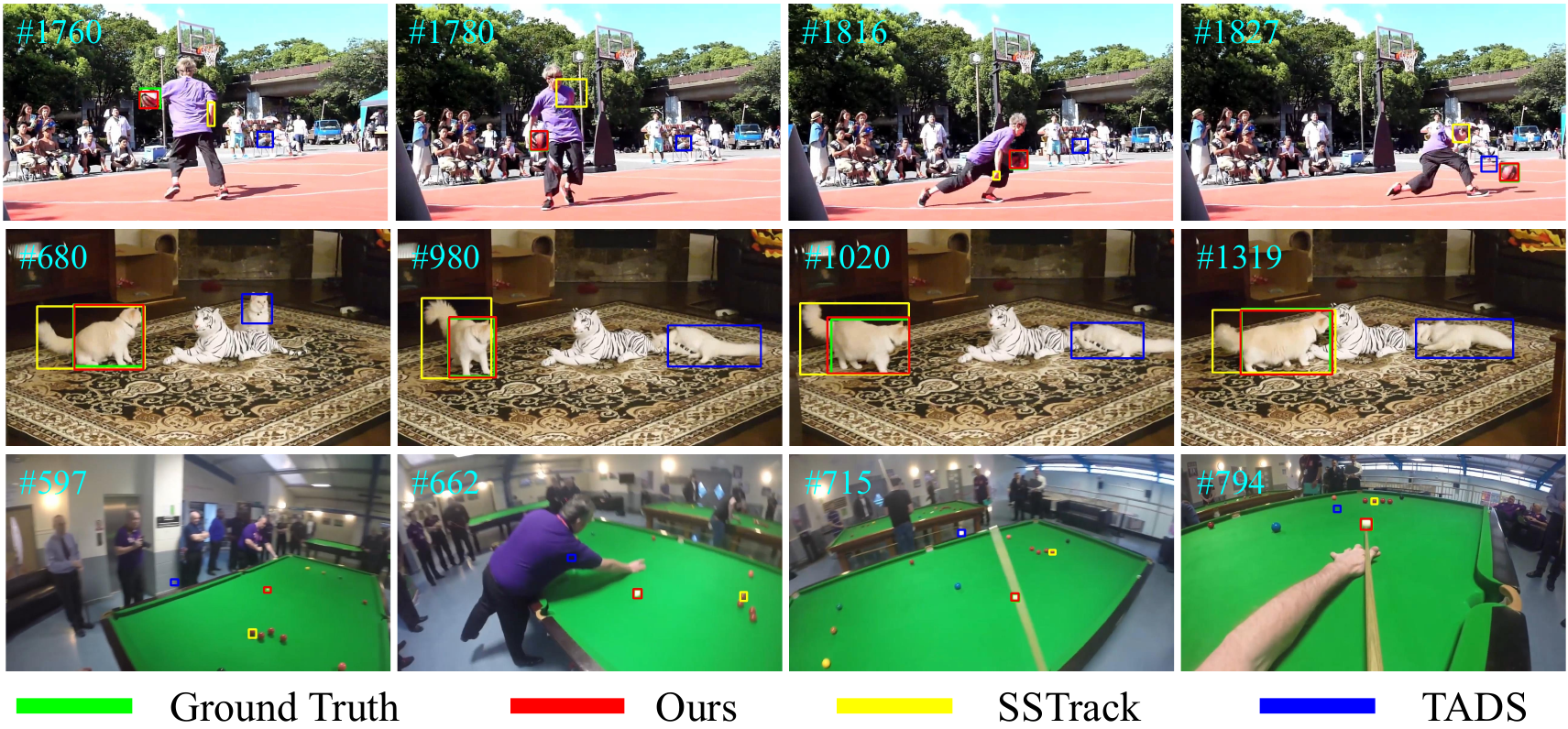}
       \caption{Qualitative comparison results of our tracker with other two SOTA trackers on LaSOT benchmark.
       }
       \label{fig:visual2}
    \end{figure}

On the other hand, we also conduct visualization experiments across multiple challenging tracking scenarios in comparison with several state-of-the-art self-supervised trackers. As illustrated in the Fig.\ref{fig:visual2}, when the billiard ball is struck and moves rapidly, our proposed tracker produces more accurate bounding boxes than the latest self-supervised tracker SSTrack. These results demonstrate that injecting contextual noise into the model facilitates the learning of more robust self-supervised features.

\section{Conclusion}
This study has proposed a self-supervised tracking framework, named {\tracker}, to efficiently learn tracking representations from unlabeled videos.
Specifically, we have introduced a novel dual-mode contextual association mechanism that adaptively modulates the model learning difficulties. Following the \textit{easy-to-hard} learning principle, we have first sampled contextual instance prompts in the early training stage to improve learning efficiency. In the later stage of training, we have injected background tokens (contextual noise) to perturb the feature space, encouraging the tracker to learn robust representations from unlabeled videos. Extensive experiments have demonstrated the superiority and effectiveness of our self-supervised tracking method.

\textbf{Limitation.}
This study has proposed a novel self-supervised tracking algorithm based on a dual-mode contextual association mechanism (DCA). Although it has achieved impressive results in the field of visual object tracking, we have not yet explored the generalizability of the DCA to other domains. Therefore, future work will extend this approach to the domain of semi-supervised video segmentation, aiming to develop a unified model for both semi-supervised video tracking and segmentation.

\textbf{Acknowledgement.} This work is supported by the Project of Guangxi Science and Technology (No. 2025GXNSFAA069676, 2024GXNSFGA010001, 2025GXNSFAA069417, and GuiKeFN2504240017), the National Natural Science Foundation of China (No.U23A20383, 62472109 and 62466051), the Guangxi ”Young Bagui Scholar” Teams for Innovation and Research Project, the Research Project of Guangxi Normal University (No. 2025DF001).
{
    \small
    \bibliographystyle{ieeenat_fullname}
    \bibliography{main}
}



\newpage

\section{Appendix}
\textbf{Additional Backgrounds.} 
With the advancement of deep learning techniques \cite{wu2025multi,wu2025spatiotemporal,ning2025multi,jiao2026large,zhang2026adaptive,zhang2026multivariate,li2025domain,ke2025detection,yu2024identifying,ReTrack,ConeSep} and the potential to eliminate the need for large-scale labeled data, self-supervised tracking has attracted increasing attention from researchers. Taking advantage of intrinsic correlations in unlabeled video data, such as temporal consistency, self-supervised tracking has shown promising results in relatively simple tracking scenarios. However, in long-term complex unlabeled tracking settings, it remains a significant challenge to capture cross-frame motion patterns and to learn robust target representations.

\textbf{Evaluation Metrics.} The tracking performance is evaluated using the toolkit corresponding to the dataset. We follow the evaluation protocol of published datasets and employ three metrics to ensure a fair comparison across various tracking methods, including success score (AUC), normalized precision score (P$_{\rm{Norm}}$), and precision score (P).

\end{document}